\documentclass[runningheads]{llncs}

\usepackage{eccv}

\usepackage{eccvabbrv}

\usepackage{graphicx}
\usepackage{booktabs}

\usepackage[accsupp]{axessibility}  % Improves PDF readability for those with disabilities.

\usepackage{hyperref}

\usepackage{orcidlink}
\usepackage{pgfplots}
\usepgfplotslibrary{groupplots}
\usepackage{amsmath}
\usepackage{xcolor}
\usepackage{multirow}
\usepackage{booktabs}
\usepackage{hyperref}
\usepackage{etoolbox}
\usepackage{makecell}
\usepackage{stfloats}
\usepackage{bbm}
\usepackage{layouts}
\usepackage{tikz}
\usetikzlibrary{shapes}
\usetikzlibrary{fit}
\usepackage{pifont}% http://ctan.org/pkg/pifont
\usepackage{algorithm}
\usepackage{algpseudocode}

\pgfplotsset{compat=1.18}

\DeclareMathOperator*{\argmax}{arg\,max}

\newcommand{\cmark}{\textcolor{olive}{\ding{51}}}%
\newcommand{\xmark}{\textcolor{red}{\ding{55}}}%
\newcommand{\aka}{\textit{a.k.a.~}}
\begin{document}

\title{CHOTA: A Higher Order Accuracy Metric for Cell Tracking} 

\titlerunning{CHOTA}

\author{Timo Kaiser\inst{1}\orcidlink{0000-0002-2745-4649} \and
Vladim{\'\i}r Ulman\inst{2}\orcidlink{0000-0002-4270-7982} \and
Bodo Rosenhahn\inst{1}\orcidlink{0000-0003-3861-1424}}

\authorrunning{T.~Kaiser et al.}

\institute{Institute for Information Processing – Leibniz University Hannover, Hannover, Germany\\
\email{\{kaiser,rosenhahn\}@tnt.uni-hannover.de} \and
IT4Innovations, VSB – Technical University of Ostrava, Ostrava, Czech Republic\\
\email{vladimir.ulman@vsb.cz}}

\maketitle

\begin{abstract}
The evaluation of cell tracking results steers the development of tracking methods, significantly impacting biomedical research. 
This is quantitatively achieved by means of evaluation metrics. 
Unfortunately, current metrics favor local correctness and weakly reward global coherence, impeding high-level biological analysis. 
To also foster global coherence, we propose the CHOTA metric (\textit{Cell-specific Higher Order Tracking Accuracy}) which unifies the evaluation of all relevant aspects of cell tracking: cell detections and local associations, global coherence, and lineage tracking. 
We achieve this by introducing a new definition of the term `trajectory' that includes the entire cell lineage and by including this into the well-established HOTA metric from general multiple object tracking. 
Furthermore, we provide a detailed survey of contemporary cell tracking metrics to compare our novel CHOTA metric and to show its advantages. 
All metrics are extensively evaluated on state-of-the-art real-data cell tracking results and synthetic results that simulate specific tracking errors. 
We show that CHOTA is sensitive to all tracking errors and gives a good indication of the biologically relevant capability of a method to reconstruct the full lineage of cells. 
It introduces a robust and comprehensive alternative to the currently used metrics in cell tracking.
Python code is available at \href{https://github.com/CellTrackingChallenge/py-ctcmetrics}{https://github.com/CellTrackingChallenge/py-ctcmetrics}.
\end{abstract}

\section{Introduction}
\label{sec:intro}
Cell tracking aims to find and cluster cell instances in time-lapse videos to model spatio-temporal cell trajectories and their lineage relations. This automates laborious work in biomedical research and allows extensive research in complex domains, \eg nervous system analysis~\cite{lovas2021ensemble} or spinal cord development~\cite{may2018cell}.

% History of cell tracking using CTC, CTMC etc. 
Unfortunately, algorithms from general multiple object tracking (MOT) are usually not easily directly applicable because cells differ in their properties. 
Different cell instances are similar in their visual appearance, have no smooth movements due to large time steps in the time-lapse, and in particular cells divide during proliferation introducing parental relations between cell instances in the lineage.   
In contrast to MOT, which has a variety of publicly available datasets (\eg \cite{milan2016mot16,dendorfer2020mot20,Geiger2012CVPR,sun2022dancetrack,abeysinghe2023tracking,xu2018youtube}), there is less publicly available data for cell tracking since imaging and labeling require expensive hardware and biological experts with domain knowledge. 
To leverage the process of algorithmic development in cell tracking, the 
ongoing \textit{Cell Tracking Challenge} (CTC)~\cite{mavska2014benchmark} was founded and provides a variety of public training data and annotations as well as undisclosed test annotations that allow a fair comparison on an evaluation server. While the CTC remains the most impacting cell tracking challenge and constantly evaluates the benchmark~\cite{ulman2017objective, celltrackingchallenge}, other public challenges~\cite{ctmc} and datasets~\cite{ker2018phase,moen2019accurate,ma2024multimodality} followed and enrich the diversity and modalities. Also, synthetic data generation pipelines~\cite{svoboda2016mitogen,sorokin2018filogen,sturm2024syncellfactory} were introduced to generate data.  

% Current metrics overview with reference to general MOT (HOTA) and that the metric drastically changes the top ranks. describe sensitivity, equality and continuity 
While the accuracy of methods increases over time, the main quality measures and metrics have not been changed. As discussed in~\cite{Eisenmann_2023_CVPR}, this phenomenon can lead to ongoing improvements in specific aspects of tracking although dismissing other aspects that, in the worst case, are relevant for biomedical practitioners. 
Current tracking metrics \cite{10.1371/journal.pone.0144959,BernardinStiefelhagen2008_1000026323} focus on local correctness like detection accuracy and correct frame-to-frame associations. 
These measures rarely reflect the global perspective on tracking that questions the ability to correctly identify an instance over long temporal periods or lineage relations between cell trajectories. 
As shown later, there are secondary metrics that evaluate the latter but do not adequately address the local errors.
Following~\cite{Chen_2023_ICCV}, an ideal metric should be \textit{sensitive} to all types of errors, \textit{equally} treat errors of the same kind, and \textit{continuously} exploit the metric's outcome interval fully to enable to rank competing methods. 
Observing that state-of-the-art methods on CTC accomplish metrics close to $100\%$ on mostly all datasets even without correct lineage reconstruction, the current main metrics seem to miss either equality, sensitivity, or continuity. 

The current cell tracking evaluation prefers local-centric methods~\cite{loeffler2022embedtrack,chen2021celltrack,loffler2021graph} and does not reward conceptual improvements of global-centric approaches~\cite{ben2022graph,Bao_2021_ICCV,nguyen2021tracking,kaiser2024cell}. 
To solve this issue in general MOT, Luiten \etal~\cite{luiten2020IJCV} introduced the HOTA metric (\textit{Higher Order Tracking Accuracy}) that holistically unifies local and global tracking correctness by measuring the impact of local decisions on global coherence. 
HOTA puts global optimal approaches in perspective and promotes the development of global approaches like~\cite{HorKai2021,Cetintas_2023_CVPR,10204724}. 
Unfortunately, HOTA does not address the biologically relevant proliferation of cells or lineage tracking. 

\begin{figure}[tb]
\centering

\begin{subfigure}{.32\textwidth}
\centering
\includegraphics[width=1\textwidth]{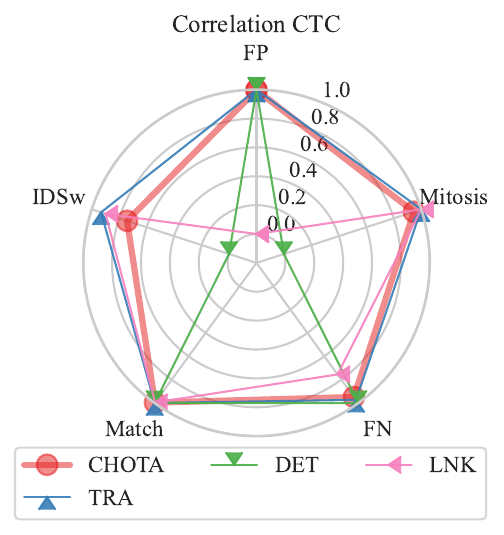}
%\caption{A}
%\label{fig:a}
\end{subfigure}
\begin{subfigure}{0.32\textwidth}
\centering
\includegraphics[width=1\textwidth]{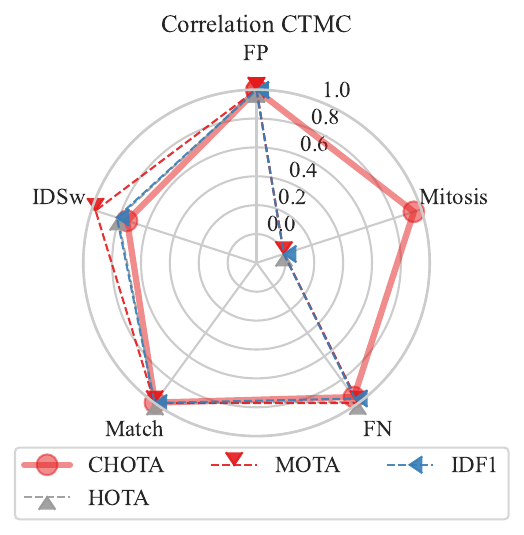}
%\caption{B}
%\label{fig:b}
\end{subfigure}
\begin{subfigure}{.32\textwidth}
\centering
\includegraphics[width=1\textwidth]{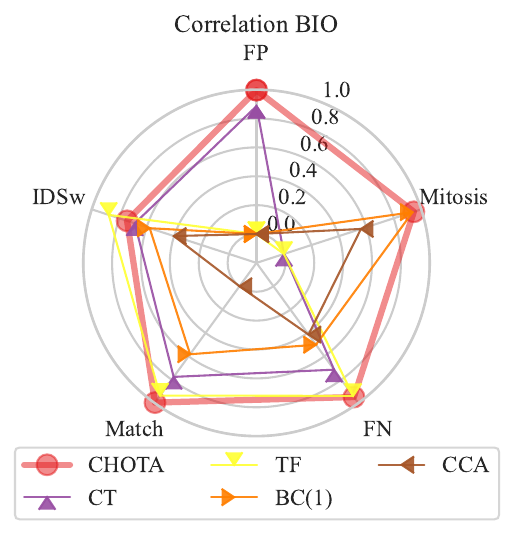}
%\caption{B}
%\label{fig:b}
\end{subfigure}

\vspace{-8px}
\caption{Sensitivity of selected cell tracking metrics to specific tracking errors, measured by the correlation in our experiments. 
Error types are identity switches (IDSw), false (FP) or missing (FN) detections, unmatched detection-annotation pairs (Match), and missed mitosis detections (Mitosis). 
From left to right, we present the CTC~\cite{celltrackingchallenge}, CTMC~\cite{ctmc} with HOTA~\cite{luiten2020IJCV}, and biological metrics (BIO) together with CHOTA.
Some metrics are less sensitive to a particular type of error (correlation is $\ll 1.0$), or even not sensitive at all ($ = 0.0$; \eg CTMC metrics to mitosis detection errors).
CHOTA and TRA are \textit{sensitive} enough to all tracking issues, but CHOTA is the only metric that also rates individual errors by their global impact. 
Thus, the same amount of IDSw (or Mitosis) errors may influence variously (depending on how much the randomly inserted errors cancel out themselves) the global coherence, which acts as noise to the correlation and is actually visible in the correlation $< 1.0$.}
\label{fig:correlation}
\end{figure}

% Overview of CHOTA and its results
To tackle these challenges, this paper proposes a metric called CHOTA (\textit{Cell-specific Higher Order Tracking Accuracy}) that brings together local correctness, global coherence, and lineage tracking, aiming to present a holistic view of the cell tracking problem. 
CHOTA builds upon HOTA but follows a novel definition of the term `trajectory' that includes lineage and thus covers biologically relevant aspects. 
We conduct extensive experiments on state-of-the-art results on nine (9) CTC datasets and specific error scenarios to reveal insights about metric behaviors. 
We analyze and compare CHOTA to all relevant metrics in the field and show that CHOTA is the only metric today that \textit{equally} treats relevant errors, is \textit{sensitive} to all aspects of cell tracking, and presents the evaluation in a \textit{continuous} domain. The experimental results are quantitatively and qualitatively summarized in Figure~\ref{fig:correlation} and Table~\ref{tab:benchmark_metrics}, respectively. 

% Overview of the paper and short results 
To compare CHOTA, this paper first offers an in-depth survey of all relevant cell tracking metrics in Section~\ref{sec:metrics}. 
Then, we introduce CHOTA in Section \ref{sec:chota} and show experiments to benchmark CHOTA and all relevant metrics in Section~\ref{sec:experiments}.

\section{Quality Measures in Cell Tracking}
\label{sec:metrics}
\begin{table*}[t]
    \centering
    \caption{Cell tracking metrics and their \textit{sensitivity} to specific error cases, the \textit{equality} to similar errors, 
    and the \textit{continuity} between 0 and 1. Furthermore, local, global, and lineage evaluation capabilities are rated. Random influences are indicated with (\cmark).}
    \resizebox{\linewidth}{!}{
        \begin{tabular}{cc|ccccccc|cccccc|c|c}
            \toprule
            % ... (table headers)
            \multicolumn{2}{c|}{} & \multicolumn{7}{c|}{CTC} & \multicolumn{6}{c|}{CTMC} \\
            \multicolumn{2}{c|}{} & TRA & DET & LNK & CT & TF & $\text{BC}(i)$ & CCA  & MOTA & IDF1 & Prec. & Rec. & MT & ML & HOTA & \textbf{CHOTA} \\
            \midrule
            
            \multirow{5}{*}{\rotatebox{90}{Sensitivity}} 
            & False Negatives & \cmark & \cmark & (\cmark) & \cmark & \cmark & (\cmark) & \cmark & \cmark & \cmark & \xmark & \cmark & \cmark & \cmark & \cmark & \cmark \\
            & False Positives & \cmark & \cmark & \xmark & \cmark & \xmark & \xmark & \cmark & \cmark & \cmark & \cmark & \xmark & \xmark & \xmark & \cmark & \cmark \\
            & ID Switches & \cmark & \xmark & \cmark & \cmark & \cmark & \xmark & (\cmark) & \cmark & \cmark & \xmark & \xmark & \cmark & \cmark & \cmark & \cmark \\
            & \ Missing Matches \ & \cmark & \xmark & \cmark & \cmark & \cmark & \xmark & \xmark & \cmark & \cmark & \cmark & \cmark & \cmark & \cmark & \cmark & \cmark \\
            & Missed Mitosis & \cmark & \xmark & \cmark & \xmark & \xmark & \cmark & \cmark & \xmark & \xmark & \xmark & \xmark & \xmark & \xmark & \xmark & \cmark \\
            \midrule
            \multicolumn{2}{c|}{Equality} & \cmark & \cmark & \cmark & \xmark & \xmark & \cmark & \xmark & \cmark & \xmark & \cmark & \cmark & \xmark & \xmark & \cmark & \cmark \\
            \multicolumn{2}{c|}{Continuity} & \xmark & \xmark & \xmark & \cmark & \cmark & \cmark & \cmark & \cmark & \cmark & \cmark & \cmark & \cmark & \cmark & \cmark & \cmark \\
            \midrule
            \multicolumn{2}{c|}{Local Correctness} & \cmark & \cmark & \cmark & \cmark & \cmark & \xmark & \xmark & \cmark & \xmark & \xmark & \xmark & (\cmark) & (\cmark) & \cmark & \cmark \\
            \multicolumn{2}{c|}{Global Coherence} & \xmark & \xmark & \xmark & \cmark & \cmark & \xmark & (\cmark) & \xmark & \cmark & \xmark & \xmark & \cmark & \cmark & \cmark & \cmark \\
            \multicolumn{2}{c|}{Lineage Tracking} & \cmark & \xmark & \cmark & \xmark & \xmark & \cmark & (\cmark) & \xmark & \xmark & \xmark & \xmark & \xmark & \xmark & \xmark & \cmark \\
            \bottomrule
        \end{tabular}
    }    
    \label{tab:benchmark_metrics}
    \vspace{-10pt}
\end{table*}

\vspace{-10px}

This section surveys evaluation protocols and quality measures of the most popular cell tracking 
benchmarks that facilitate fair evaluation conditions due to inaccessible test annotations. % and an evaluation server. 
We discuss the \textit{`Cell Tracking Challenge'}\footnote{\url{http://celltrackingchallenge.net/}} (CTC) \cite{celltrackingchallenge} and the \textit{`Cell Tracking with Mitosis Detection Dataset Challenge'}\footnote{\url{https://motchallenge.net/data/CTMC-v1/}} (CTMC) \cite{ctmc} to consolidate formal details of state-of-the-art cell tracking evaluation in one paper. Before presenting the details, preliminaries, and notations are discussed. 

All quality measures in cell tracking require a reference set of \textit{cell annotations} (\aka \textit{ground truth}) that correctly describe the real spatial and temporal position of unique cell instances in image sequences and their lineage relations. 
Usually, the ground truth is organized in spatial instance annotations labeled with IDs. The IDs are consistent over the image sequence and cluster annotations to trajectories (\aka \textit{tracks}).
% VLADO: I somehow cannot add comments to Overleaf, so explaining here:
% "also termed as" I propose only to shorten the text, as I propose to add words elsewhere
Given only raw images, the task of cell tracking is to create a set of \textit{cell predictions} (also termed as \textit{predicted detections} or just \textit{detections}) with IDs that make up tracks and establish parent-daughter relations, all of which should be equal to the ground truth. 
The quality of predictions is evaluated using metrics that overall describe the amount disparities to the ground truth, typically with a number between 0 (worst) and 1 (perfect). 

From the evaluation point of view, a distinction is made between \textit{tracking}, \textit{segmentation}, and \textit{detection}. While \textit{detection} is based on binary decisions if a cell annotation is sufficiently addressed by a corresponding cell prediction or not, \textit{segmentation} incorporates the prediction quality (typically the accuracy of cell outline). \textit{Tracking} focuses on associations of cell instances to form trajectories. 

To compare prediction and ground truth, a matching needs to be performed that assigns predicted detections to annotations and defines a set of \textit{True Positive} (TP) pairs $c\in \text{TP}$. For such a pair $c$, the ID of the corresponding detection and annotation are denoted as $\text{prID}(c)$ and $\text{gtID}(c)$, respectively.
Unmatched detections are called \textit{False Positives} (FP) and have a corresponding $\text{gtID}(c)=0$. 
Similarly, unmatched annotations are called \textit{False Negatives} (FN) for which $\text{prID}(c)=0$. 
The parent-daughter relations in the ground truth or predictions are expressed by $p(a)=b$, indicating that a cell with ID $b$ is the parent of $a$. 

%It is worth noting that
As in~\cite{luiten2020IJCV}, we use the term \textit{match} to describe the correspondence between detection and annotation and \textit{association} if objects are assigned to the same ID. 

\subsection{The \textit{`Cell Tracking Challenge'} Measures}
\label{sec:ctc_measures}

The CTC \cite{celltrackingchallenge} provides multiple datasets to evaluate tracking, detection, and segmentation. For each dataset, two training and two test sequences exist. The training data contains only tens of pixel-wise accurate ground truth annotations for segmentation (\ie not every instance is annotated) and all but simplified annotations for tracking and detection (\ie every instance is annotated with a small circle in every frame).

According to the CTC, a predicted detection, represented as a set of pixels $S_c$, is matched to a ground truth annotation, pixel set $R_c$, and denoted as a $c \in $ TP if the size of the two sets intersection is greater than half the annotation size. That way, all True Positive matches are as follows
\begin{equation}
    \text{TP} = \big\{\, \, c\, \, \big|\, \, |S_c \cap R_c| > 0.5\, |R_c|\big\}.
\end{equation}
This formulation restricts an annotation to be matched to at most one detection but allows multiple matches for a detection. 
This captures situations where a mitosis event is detected too late and both daughter cells are matched to one detection. Furthermore, the CTC does not allow temporal gaps in trajectories, leading to fragmentation (change of IDs) if a single instance is not detected. 

Following these conventions, the CTC evaluates and ranks submitted approaches in a monthly cycle based on their detection, segmentation, and tracking capabilities. The ranking is done with the following measures.

\paragraph{Segmentation Accuracy (SEG).}
Being the only segmentation metric in this paper, the SEG score evaluates predictions with the Jaccard score $J(R,S)$~\cite{jaccard1901etude} (\aka IoU). To achieve the best results, a pixel-accurate detection must be matched to each annotation. The SEG metric is defined by an averaged Jaccard score: 
\begin{equation}
    \text{SEG} = \sum_{c\in \text{TP}}\frac{ J(R_c, S_c)}{|\text{TP}| + |\text{FN}|}.
\end{equation}

\paragraph{Complete Tracks (CT).}
The CT \cite{articlect,ulman2017objective} metric describes the fraction of ground truth trajectories that are completely tracked without errors. 
A complete track has all its annotations assigned to detections with the same ID. Formally, CT is defined as the $\text{F}_1$ score over the ground truth and predicted tracks with valid IDs $i$ and $j$:
\begin{equation}
    \text{CT} = \frac{2\cdot \Big|\Big\{
    (i,j)\, \big| \, \big\{c\in \text{TP} \cup \text{FN} \,|\, \text{gtID}(c)=i \big\} \subseteq \big\{c\in \text{TP} \,|\, \text{prID}(c)=j \big\}
    \Big\}\Big|}
    {|\text{gtID}|+|\text{prID}|}
\end{equation}

\paragraph{Largest Track Fraction (TF).}
The TF \cite{ulman2017objective} metric is a relaxation of CT and averages the largest continuously tracked fraction of a ground truth track. It is important to note that TF does not mimic the $\text{F}_1$ score as CT and is therefore not sensitive to FP. The metric is defined as
\begin{equation}
    \text{TF} = \frac{1}{|\text{gtID}|} \sum_{i\in\text{gtID}}\frac{
    \max_{j\in\text{prID}}\big| \big\{c\in \text{TP} \,|\, \text{gtID}(c)=i\ \land \ \text{prID}(c)=j \big\} \big|}
    {\big| \big\{c \in \text{TP} \cup \text{FN} \,|\, \text{gtID}(c)=i \big\} \big|}.
\end{equation}

\paragraph{Cell Cycle Accuracy (CCA).}
To reflect the ability of an algorithm to discover the true distribution of cell cycle lengths in an image sequence, the CCA \cite{ulman2017objective} metric was introduced. A cell cycle is the time from birth to death/branch and is only defined for tracks that are initiated and terminated by branching events (\ie mitosis) visible in the data. The CCA measure indicates the maximal distance between the normalized cumulative distribution functions $\text{CDF}(t)$ of the predicted and ground truth cell life cycle lengths $t$. The metric is formulated as
\begin{equation}
    \text{CCA} = 1 - \max_t \big( |\text{CDF}_\text{pr}(t) - \text{CDF}_\text{gt}(t)| \big).
\end{equation}

\paragraph{Branching Correctness (\text{BC}).}
To measure the quality of mitosis detection, $\text{BC}(i)$ \cite{6957576,Bise-2011-17090,ulman2017objective} is the $\text{F}_1$ score of the detected branching events. A branching event for a match $c$ is detected if the corresponding tracks $\text{prID}(c)$ and $\text{gtID}(c)$ are both branching in the same frame or with a temporal distance up to $i$ frames. With correctly detected branching events ($\text{BTP}(i)$), undetected events ($\text{BFN}(i)$), and false detections ($\text{BFP}(i)$), the score is defined as     
\begin{equation}
    \text{BC}(i) = \frac{2\text{BTP}(i)}{2\text{BTP}(i)+\text{BFP}(i)+2\text{BFN}(i)}.
\end{equation}

\paragraph{Acyclic oriented Graph Measure (AOGM).}
The AOGM \cite{10.1371/journal.pone.0144959} metric counts all tracking and detection errors by spanning a predicted and ground truth graph and aggregating the operations that are necessary to transform the first into the latter. 
A graph is constructed by adding vertices for every detection/annotation and link-edges to connect vertices that are associated with the same track and are subsequent in time. Furthermore, parent-edges are added between the end vertex of a parent and the start vertex of the daughters. The following operations to transform the predicted graph into the ground truth graph are counted: 
Remove (FP) or add (FN) a vertex, remove (ED) or add (EA) an edge, split a vertex if a detection is falsely matched to multiple annotations (NS), and alter the semantic from link-edge to parent-edge or vice versa (EC). 
The measure is defined as the weighted sum of all errors, in which the weights reflect the effort to manually curate the tracking result and correct the respective errors:
\begin{equation}
    \text{AOGM} = w_\text{NS} \cdot \text{NS} + w_\text{FN} \cdot \text{FN} + w_\text{FP} \cdot \text{FP} + w_\text{ED} \cdot \text{ED} + w_\text{EA} \cdot \text{EA} + w_\text{EC} \cdot  \text{EC}.
\end{equation} 

\paragraph{Tracking- (TRA), Detection- (DET), and Linking  Accuracy (LNK).}
To transform AOGM into a metric, the $\text{AOGM}_0$ score is introduced that counts the operations to create the ground truth graph from scratch. The main tracking metric of the CTC is TRA and, to give values between 0 and 1, is defined by the ratio of AOGM and $\text{AOGM}_0$: 
\begin{equation}
    \text{TRA} = 1 - \frac{\min \{\text{AOGM},\text{AOGM}_0 \}}{\text{AOGM}_0}.
\end{equation} 
To measure the detection quality, DET does not incorporate linking related errors (\ie $w_\text{ED}=w_\text{EA}=w_\text{EC}=0$). Similar, LNK only measures the linking quality by removing the vertex-related errors (\ie $w_\text{NS}=w_\text{FN}=w_\text{FP}=0$).

\paragraph{Biologically relevant (BIO) and Overall Performance (OP).}
To comprehensively rank competition submissions, the CTC introduces metrics that harmonize multiple measures. The biologically relevant metrics CT, BC, TF, and CCA are averaged to 
\begin{equation}
    \text{BIO} = 0.25\cdot(\text{CT}+\text{BC}(i)+\text{TF}+\text{CCA}).
\end{equation}
To focus solely on segmentation or tracking, the DET and SEG metrics, and SEG and TRA are averaged, respectively:
\begin{equation}
    \text{OP}_\text{CSB}=0.5 \cdot(\text{DET}+\text{SEG}),  \qquad
    \text{OP}_\text{CTB}=0.5 \cdot(\text{DET}+\text{TRA}).
\end{equation}
Recently, the $\text{OP}_\text{CLB}$ metric was added to measure the overall performance of association strategies. 
% showing that we follow the field (with "Recently added"), and saving space.
To remove the influence of the detection quality, $\text{OP}_\text{CLB}$ is evaluated on a prepared set of detections and is defined as
\begin{equation}
    \text{OP}_\text{CLB}=0.5\cdot(\text{BIO}+\text{LNK}). 
\end{equation}

\subsection{The \textit{`CTMC-MOTChallenge'} Measures}
\label{sec:ctmc_measures}
The CTMC \cite{ctmc} provides a large-scale cell tracking dataset covering 14 cell types. Compared to the CTC, CTMC includes fewer tracks (2900 vs.~11318) but has more frames (152584 vs.~5927). Annotations for cell instances are given as bounding boxes that cause ambiguous associations of pixels to cell instances because a pixel can be included in multiple bounding boxes. 
The matching procedure between detections and annotations depends on the metric. 
The CTC matching is used to calculate the TRA metric. On the other hand, Hungarian matching \cite{Kuhn1955Hungarian} creates a bijective mapping between detections and annotations. The costs for the matching are created using the Jaccard score: 
\begin{equation}
    \argmax_{\text{TP}} \sum_{c\in \text{TP}} J(R_c, S_c)
    \quad \text{with} \quad \forall c \in \text{TP}:\, J(R_c, S_c) > 0.5
\end{equation} 
A match is typically only allowed, iff the Jaccard score is larger than a threshold, usually $0.5$. This strategy does not allow multiple matches to a detection.

Since there are no pixel-wise segmentation masks, the CTMC does not provide a segmentation benchmark. Submitted methods are evaluated with metrics that are sensitive to tracking and detection errors, jointly. 
The CTMC evaluates submitted results ad-hoc, but evaluations per month is limited to avoid test-data fitting. The next section presents metrics used by the CTMC. 
Furthermore, FP, FN, and TRA scores introduced in Section~\ref{sec:ctc_measures} are also reported by the CTMC.

\paragraph{Multiple Object Tracking Accuracy (MOTA) and Identity Switches (IDSw).}
The main CTMC tracking metric to rank approaches is the MOTA metric \cite{BernardinStiefelhagen2008_1000026323}. 
Using IDSw that denote association errors, in which two subsequent detections $S_{c_1}$ and $S_{c_2}$ of a predicted trajectory (\ie $\text{prID}(c_1)=\text{prID}(c_2)$) are matched to different ground truth trajectories (\ie $\text{gtID}(c_1)\neq\text{gtID}(c_2)$), MOTA is defined as:
\begin{equation}
\label{eq:mota}
    \text{MOTA} = 1 - \frac{|\text{FN}|+|\text{FP}| + |\text{IDSw}| }{|\text{TP}|+|\text{FN}|}.
\end{equation}

\paragraph{Identification $\text{F}_1$ Score ($\text{IDF}_1$).}
The $\text{IDF}_1$ metric \cite{ristani2016performance} measures the $\text{F}_1$ score with TPs, FPs, and FNs but only considers TPs that belong to matched trajectory pairs. The metric assumes that a ground truth trajectory $i$ can only be assigned to a single predicted trajectory $j$, and vice-versa. Matching trajectory pairs are defined by an optimized trajectory assignment $(i,j)\in M$, such that the $\text{IDF}_1$ score is maximized. The overall metric is defined as
\begin{equation}
    \text{IDF}_1 = \max_M \Bigg(
    \frac{\sum_{(i,j)\in M}
    2\cdot \big| \big\{c\in \text{TP} \,|\, \text{gtID}(c)=i\ \land \ \text{prID}(c)=j \big\} \big|}
    {2\big|\text{TP}\big|+\big|\text{FN}\big|+\big|\text{FP}\big|}
    \Bigg).
\end{equation}

\paragraph{Recall, Precision, and False Alarms per Frame (FAF).}
The detection quality is also evaluated using recall, precision, and the FAF quantity. The FAF quantity is the number of FP normalized over the total number of frames in the image sequence. With an image sequence $\mathcal{I}$, the measures are defined as 
\begin{equation}
    \text{Precision}= \frac{|\text{TP}|}{|\text{TP}| + |\text{FP}|},  \quad  \text{Recall}= \frac{|\text{TP}|}{|\text{TP}| + |\text{FN}|}, \quad \text{and} \quad \text{FAF}=\frac{|\text{FP}|}{|\mathcal{I}|}.
\end{equation}

\paragraph{Mostly Tracked (MT) and Mostly Lost (ML).}
The CTMC provides the additional criteria MT and ML to evaluate the tracking performance. The measure MT is the ratio of ground-truth trajectories that are covered by a track hypothesis for at least $80\%$ of their respective life span. ML indicates the ratio with at most $20\%$ coverage, respectively. 

\section{Unifying Local Correctness with Global Consistency and Lineage Tracking}
\label{sec:chota}

Current cell tracking metrics do not follow ideal metric criteria~\cite{Chen_2023_ICCV}: No metric is \textit{sensitive} to all error types, \textit{equal} to similar errors, and \textit{continuous}. 
The main metrics analyze local detection quality and the association between subsequent frames (TRA, DET, LNK, MOTA), while ignoring global coherence. %, while ignoring lineage reconstruction or capabilities like the ability to reverse ID switches to form coherent trajectories. 
Secondary metrics evaluate biological characteristics and global coherence, \ie large trajectory fractions are predicted correctly (CT, TF, CCA, BC, $\text{IDF}_1$, MT, ML), but do not adequately address local correctness. 

The main focus on locality led to outstanding results in metrics like TRA while simultaneously being bad in biologically relevant metrics. 
This disparity between local correctness and global coherence was dissolved in general MOT by introducing the HOTA metric \cite{luiten2020IJCV} that unifies MOTA and $\text{IDF}_1$:
\begin{align}
   \text{HOTA} = \sqrt{\frac{\sum_{c \in \text{TP}} \mathcal{A}(c) }{|\text{TP}| + |\text{FN}| + |\text{FP}|}} 
   \label{equ:hotaalpha}
\end{align}
HOTA treats local errors similar to MOTA but weights each TP individually by its global coherence % of its trajectory 
using an association score $\mathcal{A}(c)\in[0,1]$ that is defined later. % that will be discussed later. 

However, HOTA does not address biologically relevant aspects of cell tracking, namely the cell divisions. 
To close this gap, we propose CHOTA which unifies HOTA and lineage tracking. 
In the next sections, we reinvent the term `trajectory' to include lineage information. Then, we redefine the score $\mathcal{A}(c)$ utilizing our trajectory definition. Both combined result in CHOTA. 

\begin{figure}[!tb]
\centering
\begin{subfigure}{.5\textwidth}
  \centering
  \includegraphics[trim=0 75 0 0, clip, width=1.0\linewidth]{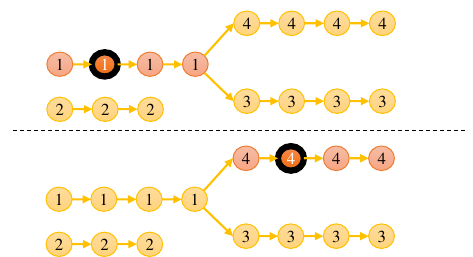}
  \caption{ID-oriented trajectory}
  \label{fig:trajectory_hota}
\end{subfigure}%
\begin{subfigure}{.5\textwidth}
  \centering
  \includegraphics[trim=0 75 0 0, clip, width=1.0\linewidth]{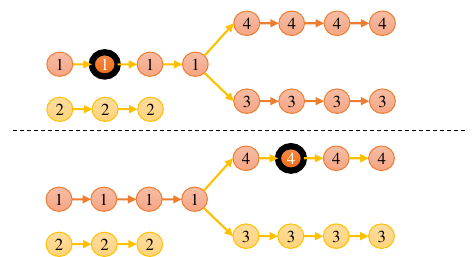}
  \caption{Lineage-oriented trajectory}
  \label{fig:trajectory_chota}
\end{subfigure}
\vspace{-10pt}
\caption{The common ID-oriented and our lineage-oriented trajectory. A node represents a cell in a specific frame with its respective ID and an edge emphasizes the evolution over time. Each line shows the same lineage tree but highlights a different trajectory (red) for a specific cell detection (black). Our definition accounts for ID-oriented information and also biologically relevant relations to ancestors and descendants.}
\vspace{-10pt}
\label{fig:trajectory}
\end{figure}

\subsection{Lineage-oriented Trajectories}

Tracking metrics evaluate the correctness of predicted trajectories. Across all (cell) tracking literature, a trajectory is defined as a set of objects that share the same ID. 
Especially in general MOT, this is reasonable and implies that objects have no relation to objects with different IDs, as indicated with the red trajectories in Figure~\ref{fig:trajectory_hota}. 
However, this does not hold for cell tracking where cells can proliferate and split into multiple daughter cells. 
Research like embryonic studies~\cite{https://doi.org/10.15252/embj.2022113280,ichikawa2022ex,10.1242/dev.082586} not only require the information if two cells have the same ID, they also need to know if they have lineage relations. Thus, an ideal cell tracking metric should evaluate trajectories that include both aspects. 

To facilitate lineage information, we redefine the notion of the term `trajectory':
Two cells with IDs $i$ and $j$ belong to the same trajectory \textbf{if} $i$ is equal to $j$ \textbf{or if} $i$ is an ancestor of $j$ or vice versa. 
The implications are visualized in Figure~\ref{fig:trajectory_chota} where the parent cell trajectory includes both daughters but the daughter trajectory has no relation to its sibling. 
This reflects that a mutation in ID 1 affects IDs 3 and 4, but ID 4 cannot be affected by ID 3. 
The trajectory can be formalized using an indicator function $\sigma(i,j)$ 
and a parent-daughter relation $\text{p}^n(a)=b$ which states that cell $b$ is the $n$-th degree ancestor of cell $a$: 
\begin{equation}
\sigma(i,j)=
\begin{cases}
1, \,  & \text{if} \,\ i=j \\ %\vee i=\text{p}(j) \vee \text{p}(i)=j \\
%1, \,  & \text{if} \,\ \exists \ i, l_1, \ldots, l_n, j \in \text{IDs} : i = \text{p}(l_1) \wedge \ldots \wedge l_n= \text{p}(j) \\
1, \,  & \text{if} \,\ \exists  n : \,\ i = p^n(j) \,\ \text{or} \,\ j = p^n(i) \\
% 1, \,  & \text{if} \,\ \exists \ j, l_n, \ldots, l_1, i \in \text{IDs} :  
% \begin{split}  A+B+C+ \\ +D+E+F \end{split}) \\
0, \,  & \text{else.}
\end{cases}
\label{eq:indicator}
\end{equation}

Considering two True Positive detections $c_1 \in \text{TP}$ and $c_2\in \text{TP}$, the lineage-oriented trajectory indicator can be used to evaluate all relevant cell tracking aspects: 
By comparing the prediction $\sigma(\text{prID}(c_1), \text{prID}(c_2))$ and ground truth $\sigma(\text{gtID}(c_1), \text{gtID}(c_2))$, local association correctness can be evaluated if $c_1$ and $c_2$ are in subsequent frames, global coherence if they have a temporal gap, and lineage relations if they do not share the same ID.

\subsection{CHOTA}
\begin{figure}[!tb]
\centering
\includegraphics[width=1.0\linewidth]{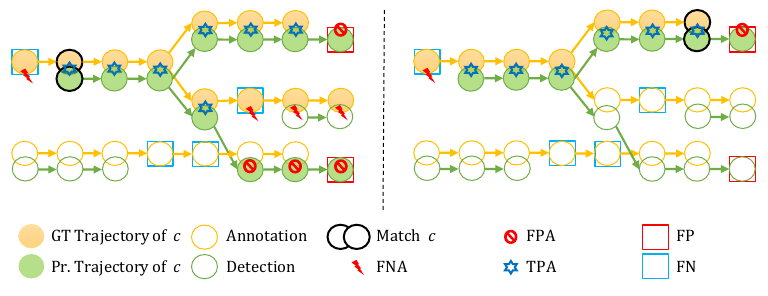}
\vspace{-20pt}
\caption{An example prediction (green) to annotation (yellow) mapping presenting the relations to calculate $\mathcal{A}^\sigma(c)$ from Equation~\eqref{eq:assoc} for two matched pairs $c$ (black). Using our lineage-oriented trajectory, TPA, FPA, and FNA reflect inter- and intra-ID relations. On the left, the ground truth annotation of $c$ relates to two daughter cells of which only one is represented in the prediction, leading to a low $\mathcal{A}^\sigma(c)=\frac{7}{15}$. On the right, the lineage of $c$ is almost correct, leading to a higher $\mathcal{A}^\sigma(c)=\frac{6}{8}$.}
\vspace{-10pt}
\label{fig:chota}
\end{figure}

The score $\mathcal{A}(c)$ from Equation~\eqref{equ:hotaalpha} evaluates the global trajectory coherence of a True Positive $c\in \text{TP}$ individually. 
This is done by comparing the corresponding predicted and ground truth trajectories with the Jaccard score~\cite{jaccard1901etude}. Qualitatively speaking, the score is high if both trajectories are similar over time. 
To open $\mathcal{A}(c)$ to our trajectory definition, we utilize our formerly introduced indicator $\sigma(i,j)$ and extend the association measures from~\cite{luiten2020IJCV} named \textit{True Positive Associations} (TPA), \textit{False Negative Associations} (FNA), and \textit{False Positive Associations} (FPA), to calculate a lineage-oriented score 
\begin{equation}
    \mathcal{A}^\sigma(c) = \frac{|\text{TPA}^\sigma(c)|}{|\text{TPA}^\sigma(c)| + |\text{FNA}^\sigma(c)| + |\text{FPA}^\sigma(c)|}.
    \label{eq:assoc} 
\end{equation}
Using the indicator, the set $\text{TPA}^\sigma(c)$ consists of TPs that are associated with the same ground truth and predicted trajectory: 
\begin{align}
\text{TPA}^\sigma(c) = \big\{ k\in\text{TP} \, \big| \, \sigma\big(\text{prID}(k), \text{prID}(c)\big) \land \sigma\big(\text{gtID}(k), \text{gtID}(c)\big) \big\}.
\label{eq:TPA}
\end{align}
They are indicated with blue stars in Figure~\ref{fig:chota} and reflect the overlap of the two trajectories, including the entire reconstructed lineage tree.
The set $\text{FNA}^\sigma(c)$ contains all annotations that are associated with the same ground truth trajectory but matched with a different or no predicted trajectory:
\begin{align}
\begin{aligned}
\multirow{2}{*}{ $\text{FNA}^\sigma(c) = $ }  & \big\{ k\in\text{TP} \, \big| \, \overline{\sigma\big(\text{prID}(k), \text{prID}(c)\big)} \land \sigma\big(\text{gtID}(k), \text{gtID}(c)\big) \big\} \\ & \quad \cup \,  \big\{ k\in\text{FN} \, \big| \, \sigma\big(\text{gtID}(k), \text{gtID}(c)\big) \big\}.
\end{aligned}
\label{eq:FNA}
\end{align}
Similarly, $\text{FPA}^\sigma(c)$ is the union of detections associated with the same predicted trajectory but matched with a different or no ground truth trajectory:
\begin{align}
\begin{aligned}
\multirow{2}{*}{ $\text{FPA}^\sigma(c) = $ }  & \big\{ k\in\text{TP} \, \big| \, \sigma\big(\text{prID}(k), \text{prID}(c)\big) \land \overline{\sigma\big(\text{gtID}(k), \text{gtID}(c)\big)} \big\} \\ & \quad \cup \,  \big\{ k\in\text{FP} \, \big| \, \sigma\big(\text{prID}(k), \text{prID}(c)\big) \big\}.
\end{aligned}
\label{eq:FPA}
\end{align}
Indicated with red markers in  Figure~\ref{fig:chota}, FNAs and FPAs count the missing and incorrect parts of the predicted trajectory, respectively.

By applying the new trajectory definition to $\mathcal{A}^\sigma(c)$, CHOTA is defined as 
\begin{align}
   \text{CHOTA} = \sqrt{\frac{\sum_{c \in \text{TP}} \mathcal{A}^\sigma(c) }{|\text{TP}| + |\text{FN}| + |\text{FP}|}} 
   \label{equ:chotaalpha}
\end{align}

The new lineage-oriented trajectory allows $\mathcal{A}^\sigma(c)$ to reward or penalize all relevant cell tracking events. 
Similar to HOTA and MOTA, CHOTA penalizes local errors like FPs and FNs in its denominator. 
Extending $\text{TPA}$, $\text{FPA}$, and $\text{FNA}$ \textit{equally} to all detections that have lineage relations to $c$ makes the reward $\mathcal{A}^\sigma(c)$ also \textit{sensitive} to lineage errors. 
The TPs are individually rewarded relative to their global trajectory coherence and thus implicitly incorporate ID switches and mitosis detections. 
The individual $\mathcal{A}^\sigma(c)$ is conditioned on the broader impact as shown in Figure~\ref{fig:chota}: 
While not traversing the coherence in the right trajectory, the wrong association in one daughter cell gets penalized in the left trajectory because it mistakenly creates a lineage relation to an independent cell instance. 

It is worth noting that CHOTA can be calculated efficiently in a quasi-linear dependency on the number of IDs as shown in the Appendix. 
Equivalent to HOTA~\cite{luiten2020IJCV}, CHOTA can be decomposed into the detection (DetA) and association (AssA) sub-metrics which are not further elaborated in this paper. 
Furthermore, we use the CTC matching procedure from Section \ref{sec:ctc_measures}. Other strategies like bijective matching from Section~\ref{sec:ctmc_measures} or HOTAs $\alpha$-integral optimization~\cite{luiten2020IJCV} can be applied, too, but require accurate segmentation annotations that are not always available, \eg in CTC.

\begin{table*}[t]
    \centering
    \caption{
    Benchmark on all tracking measures used by CTC~\cite{celltrackingchallenge} and CTMC~\cite{ctmc}, as well as HOTA~\cite{luiten2020IJCV} and our CHOTA metric, calculated on the state-of-the-art results of \textit{EmbedTrack}~\cite{loeffler2022embedtrack}. CHOTA unifies both, \textcolor{red}{general} and \textcolor{LimeGreen}{biological} tracking aspects.
    }
    \vspace{-5pt}

    % Introduce a new counter for counting the nodes needed for circling
    \newcounter{nodecount}
    % Command for making a new node and naming it according to the nodecount     counter
    \newcommand\tabnode[1]{\addtocounter{nodecount}{1} \tikz \node  (\arabic{nodecount}) {\hspace{-4.0pt} #1 };}
    
    % Some options common to all the nodes and paths
    \tikzstyle{every picture}+=[remember picture,baseline]
    \tikzstyle{every node}+=[anchor=base, outer sep=1.5pt, inner xsep = 0.0pt, inner ysep = 0.5pt, align=left, very thick]
    \tikzstyle{every path}+=[thick, rounded corners]

    \resizebox{\linewidth}{!}{
        \begin{tabular}{cc|ccccccccc}
            \toprule
            % ... (table headers)
            \multicolumn{2}{c}{} & \makecell{BF-C2DL-\\HSC} & \makecell{BF-C2DL-\\MuSC} & \makecell{DIC-C2DH-\\HeLa} & \makecell{Fluo-C2DL-\\MSC} & \makecell{Fluo-N2DH-\\GOWT1} & \makecell{Fluo-N2DL-\\HeLa} & \makecell{PhC-C2DH-\\U373} & \makecell{PhC-C2DL-\\PSC} & \makecell{Fluo-N2DH-\\SIM+} \\
            \midrule
            
            \multirow{5}{*}{\rotatebox{90}{Properties}} 
            & Frames & 3526 & 2750 & 166 & 94 & 182 & 182 & 228 & 598 & 213 \\
            & Instances & 73178 & 12706 & 2150 & 616 & 4575 & 34059 & 1457 & 128207 & 5978 \\
            & Branches & 168 & 44 & 13 & 0 & 6 & 303 & 0 & 992 & 72 \\
            & Tracks & 340 & 119 & 70 & 25 & 86 & 939 & 20 & 2395 & 202 \\
            & \ \ \ Cell Cycles\ \ \ & 165 & 39 & 0 & 0 & 0 & 151 & 0 & 783 & 36 \\
            \midrule
            
            \multirow{12}{*}{\rotatebox{90}{CTC-Metrics}} 
            & SEG $\uparrow$& 0.883 & 0.778 & 0.889 & 0.643 & 0.942 & 0.884 & 0.901 & 0.766 & 0.837 \\
            & CT $\uparrow$& \tabnode{0.032} & 0.017 & 0.329 & 0.020 & 0.529 & 0.439 & 0.251 & 0.194 & 0.574 \\
            & TF $\uparrow$& 0.721 & 0.596 & 0.857 & 0.692 & 0.960 & 0.933 & 0.920 & 0.871 & 0.923 \\
            & BC(1) $\uparrow$& 0.451 & 0.163 & 0.493 & - & 0 & 0.625 & - & 0.513 & 0.827 \\
            & CCA $\uparrow$& {0.108} & 0.051 & - & - & - & 0.615 & - & 0.498 & 0.628 \\
            & BIO(1) $\uparrow$& \tabnode{0.328} & 0.207 & 0.560 & 0.356 & 0.496 & 0.653 & 0.585 & 0.519 & 0.738 \\
            & TRA $\uparrow$& 0.976 & 0.957 & 0.952 & 0.908 & 0.977 & 0.982 & 0.960 & 0.962 & 0.973 \\
            & DET $\uparrow$& 0.974 & 0.962 & 0.953 & 0.914 & 0.978 & 0.983 & 0.963 & 0.963 & 0.975 \\
            & LNK $\uparrow$& 0.988 & 0.925 & 0.945 & 0.866 & 0.976 & 0.977 & 0.938 & 0.950 & 0.957 \\
            & $\text{OP}_\text{CSB}$  $\uparrow$& 0.929 & 0.870 & 0.921 & 0.778 & 0.960 & 0.934 & 0.932 & 0.865 & 0.906 \\
            & $\text{OP}_\text{CTB}$  $\uparrow$& 0.930 & 0.867 & 0.921 & 0.775 & 0.960 & 0.933 & 0.930 & 0.864 & 0.905 \\
            & $\text{OP}_\text{CLB}$  $\uparrow$& 0.658 & 0.566 & 0.753 & 0.611 & 0.736 & 0.815 & 0.762 & 0.735 & 0.848 \\
            \midrule

            \multirow{12}{*}{\rotatebox{90}{CTMC-Metrics}} 
            & TP $\uparrow$& 36583 & 6321 & 1039 & 299 & 2241 & 16976 & 727 & 63538 & 2925 \\
            & FP $\downarrow$& 3617 & 1705 & 137 & 163 & 81 & 2221 & 161 & 14562 & 46 \\
            & FN $\downarrow$& 6 & 32 & 36 & 9 & 46 & 53 & 1 & 565 & 63 \\
            & IDSW $\downarrow$& 421 & 335 & 10 & 25 & 5 & 199 & 1 & 1054 & 23 \\
            & Precision $\uparrow$& 0.819 & 0.783 & 0.887 & 0.632 & 0.964 & 0.890 & 0.821 & 0.814 & 0.985 \\
            & Recall $\uparrow$& 1.000 & 0.995 & 0.966 & 0.973 & 0.981 & 0.997 & 0.998 & 0.991 & 0.979 \\
            & FAF $\downarrow$& 2.1 & 1.3 & 1.6 & 3.4 & 0.9 & 24.7 & 1.6 & 50.8 & 0.6 \\
            & MT $\uparrow$& 0.428 & 0.289 & 0.810 & 0.383 & 0.934 & 0.897 & 0.929 & 0.793 & 0.879 \\
            & ML $\downarrow$& 0.002 & 0.011 & 0 & 0 & 0 & 0.001 & 0 & 0 & 0 \\
            & MOTA $\uparrow$& \tabnode{0.736} & {0.649} & 0.830 & 0.284 & 0.942 & 0.860 & 0.751 & 0.738 & 0.952 \\
            & IDF1 $\uparrow$& \tabnode{0.599} & {0.529} & 0.843 & 0.535 & 0.962 & 0.836 & 0.829 & 0.785 & 0.943 \\
            \midrule\midrule
            & HOTA & \tabnode{0.688} & {0.617} & 0.851 & 0.576 & 0.959 & 0.863 & 0.843 & 0.809 & 0.944 \\
            & \textbf{CHOTA} & \tabnode{0.542} & {0.451} & 0.849 & 0.630 & 0.947 & 0.865 & 0.859 & 0.745 & 0.959 \\
            \bottomrule
        \end{tabular}

        \begin{tikzpicture}[overlay]
        % Define the circle paths
        \node[draw=LimeGreen,anchor=center, rounded corners = 1ex,fit=(1)(2), xshift=-2pt] (bio) {};
        \node[draw=red,anchor=center, rounded corners = 1ex,fit=(3)(4), xshift=-2pt] (general) {};
        \node[draw=black,anchor=center, rounded corners = 1ex,fit=(5)(5), xshift=-2pt] (hota) {};
        \node[draw=black,anchor=center, rounded corners = 1ex,fit=(6)(6), xshift=-2pt] (chota) {};

        \path[draw=LimeGreen,-, very thick] 
            (bio.east) -- +(1em, 0) |-  (chota.east);
        \path[draw=red,-, very thick] 
            (general.east) -- +(0.5em, 0) |-  (chota.east);
        \path[draw=red,-, very thick] 
            (general.west) -- +(-0.5em, 0) |-  (hota.west);
        
        \end{tikzpicture}

    }
    
    \label{tab:benchmark_realworld}
    \vspace{-10pt}
\end{table*}

\section{Experiments and Discussion}
\label{sec:experiments}

This section presents insights about CHOTA and the metrics introduced in Section~\ref{sec:metrics}. % when errors are present in predicted tracking results. 
First, a benchmark of all metrics is created on real-world tracking results showing that CHOTA unifies the outcome of other metrics and is the most holistic measure.  
Then, specific error types are induced synthetically into perfect tracking results to analyze the influence of typical tracking failure cases on the metrics. 
The experimental results are qualitatively summarized in Table~\ref{tab:benchmark_metrics}.

\subsection{Real-World Tracking}
\label{sec:experiments_realworld}
To evaluate metric behaviors in real tracking scenarios, we re-evaluated all metrics on the state-of-the-art method \textit{EmbedTrack}~\cite{loeffler2022embedtrack}. 
To this end, we applied and evaluated the publicly available framework on the training data of 9 diverse datasets from the CTC. The selection of datasets contains various scenarios with long, short, dense, and sparsely populated image sequences with weak to heavy proliferation. 
The results and relevant dataset properties are given in Table~\ref{tab:benchmark_realworld}.

The biologically relevant metrics that reflect global tracking and lineage consistency are very small (\eg $<5\%$) on long datasets with large proliferation (BF-C2DL-HSC/MuSC) and relatively high on small datasets (\eg Fluo-N2DH-SIM+). 
Simultaneously, the more relevant and local-centric CTC main tracking metrics TRA, DET and LNK, are consistently close to 1 in mostly all results. This demonstrates the imbalance in cell tracking algorithm development and mistakenly could lead to the belief that the tracking task is close to being solved. 

Taking a closer look at the CTMC metrics MOTA or IDF1 reveals that they utilize a larger value range. Low MOTA scores indicate a large number of local errors (FP, FN, and IDSw), and low IDF1 shows that these local errors are not recovered globally over time. As shown in general MOT, the HOTA metric harmonizes local and global error measures~\cite{luiten2020IJCV}. The range of values is from $~58\%$ to $~96\%$ and indicates that the general tracking task is between half and close to being solved, leaving much space for algorithmic improvements. 
While MOTA, IDF1, and HOTA are sensitive to local, global, or both tracking aspects, they are insensitive to lineage tracking by definition and discard valuable information.  

Our proposed lineage-oriented CHOTA metric is also sensitive to biologically relevant lineage information. 
While CHOTA is close to HOTA on the small datasets where lineage information is weak, it decreases drastically on datasets with difficult trajectories where lineage construction fails according to the biological metrics (BF-C2DL-HSC/MuSC, and PhC-C2DL-PSC). 
Furthermore, CHOTA is slightly higher than HOTA on Fluo-C2DL-MSC. The cause is revealed when having a closer look into the results: The ground truth contains no mitotic events but the tracking results falsely detected over-segmentations as mitosis, \ie two detections (TP and FP) are matched to a single annotation. While this destroys an entire trajectory in HOTA, biological relations still hold in the novel lineage-oriented trajectory of CHOTA leading to the better evaluation.  
This showcases the ability of CHOTA to be sensitive to all cell tracking aspects and to provide a holistic evaluation. Moreover, it continuously employs a larger value range that helps to interpret the current state of different datasets.

\begin{figure}[!t]
\centering

\begin{subfigure}{.32\textwidth}
\centering
\includegraphics[width=1\textwidth]{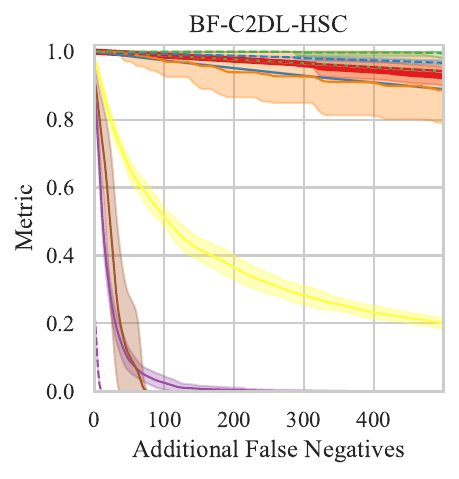}
\end{subfigure}
\begin{subfigure}{0.32\textwidth}
\centering
\includegraphics[width=1\textwidth]{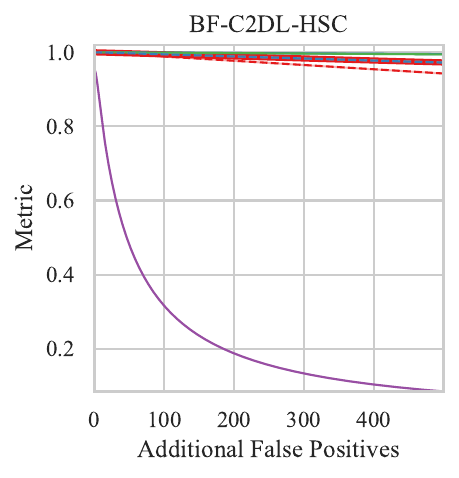}
\end{subfigure}
\begin{subfigure}{.32\textwidth}
\centering
\includegraphics[width=1\textwidth]{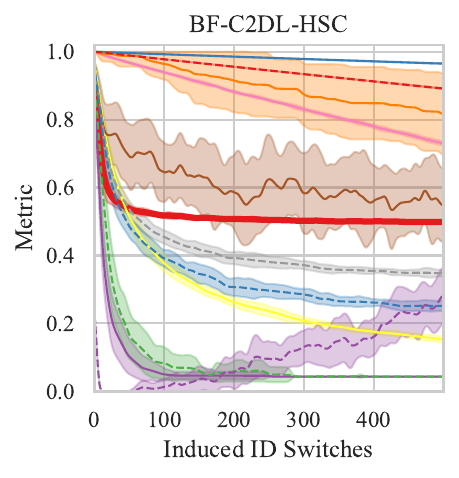}
\end{subfigure}

\begin{subfigure}{.32\textwidth}
\centering
\includegraphics[width=1\textwidth]{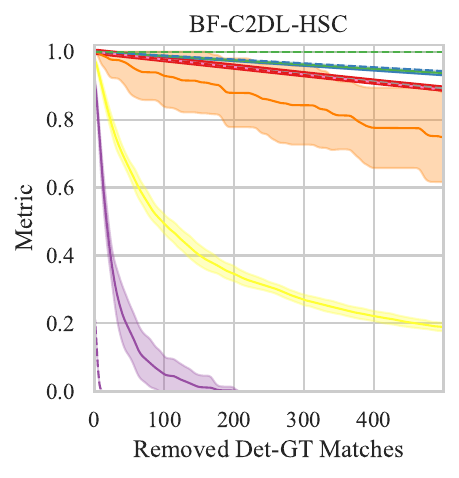}
\end{subfigure}
\begin{subfigure}{0.32\textwidth}
\centering
\includegraphics[width=1\textwidth]{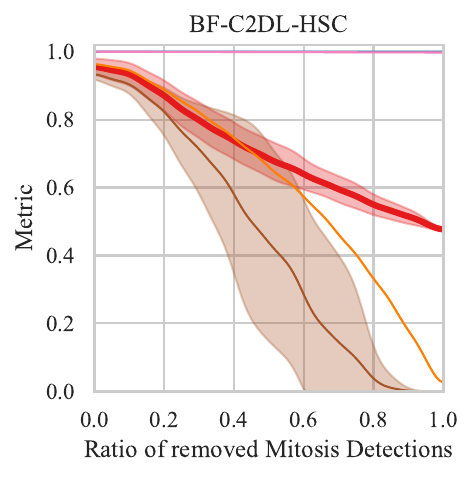}
\end{subfigure}
\begin{subfigure}{.32\textwidth}
\centering
\includegraphics[width=.4\textwidth]{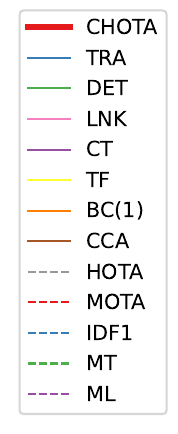}
\end{subfigure}
\caption{The behavior of metrics if specific tracking errors are induced into a perfect tracking result (BF-C2DL-HSC, Seq. 01). The biological metrics (CT, TF, BC, CCA) are not \textit{equally} treating similar errors, visible in the large variances. TRA is not adequately reflecting the main tracking issues of ID switches and missed mitosis detections, apparent in consistent values close to 1. CHOTA is the only metric that is both, \textit{equal} and \textit{continuously} exploiting a large value range when tracking errors are induced.}
\label{fig:noise_overview}
\end{figure}

\subsection{Random Tracking Error Induction}
\label{sec:experiments_synthetic}
We evaluate the influence of specific error sources to perfect tracking results. To do so, we use the ground truth as tracking result and randomly induce errors. We randomly add False Positive detections (FP), remove detections (FN), remove correct detection-to-ground truth matchings, remove parent-daughter relations (mitosis detections), and induce ID switches. Only errors of one type are added to the result at the same time and every experiment is repeated at least 10 times with different seeds. 
We use the same datasets as in Section~\ref{sec:experiments_realworld}
and visualize the smoothed mean and variance of the metrics on BF-C2DL-HSC in Figure~\ref{fig:noise_overview}. The results on all other datasets are comparable and can be found in the Appendix. 

In Figure~\ref{fig:noise_overview}, variance reflects random influence, meaning that similar errors are not treated \textit{equally}. It shows that MT, ML, CT, TF, CCA, and BC$(i)$ are prone to local randomness, while local metrics like TRA, DET, LNK, and MOTA have no variance by their definition. IDF1, HOTA, and CHOTA only incorporate small randomness on ID switches and mitosis. The randomness is caused by errors that dissolve former errors and lead to more global coherence. 

Our CHOTA metric is the only metric that is adequately \textit{sensitive} and \textit{continuous} to general tracking and mitosis detection errors. While TRA and LNK remain close to $100\%$ even if all (!) mitosis detections are removed, CHOTA decreases to $50\%$, showing the harmonization of general and lineage tracking. Although the BC$(i)$ metric is \textit{sensitive} and \textit{continuous} to mitosis detection by its definition, it is just randomly influenced by all other tracking errors.  

To summarize the effect of specific error types on the metrics, we visualize the linear correlation of the number of induced errors to the metrics in Figure~\ref{fig:correlation}. For visualization purposes, we present the magnitude of the correlation averaged over all datasets. Our CHOTA metric covers all cell tracking relevant aspects with a relatively large correlation. The only metric that also correlates to all error sources is the TRA metric. The correlation to ID switches is larger than by CHOTA which reflects CHOTAs ability to reward coherence-promoting ID switches. With few exceptions like CT to FP and BC$(i)$ to Mitosis, there is a low correlation between biologically relevant metrics and cell tracking errors, underlining the random influence in the evaluation. Overall, CHOTA is the only metric that addresses all cell tracking failure cases and global tracking coherence.

\section{Conclusion}
\label{sec:conclusion}
This work addresses the problem of unbalanced evaluation metrics in cell tracking that mostly focus on local correctness and do not appropriately reflect global coherence and lineage tracking.   
To tackle this problem, we introduce the novel metric CHOTA - the \textit{Cell-specific Higher Order Tracking Accuracy} - that unifies local and global tracking aspects as well as biologically relevant lineage consistency.
Furthermore, we provide an in-depth survey and analysis of all relevant cell tracking metrics that help researchers in their choice of metrics. 
Our analysis shows that CHOTA is the only metric that is \textit{sensitive} and \textit{equal} to all relevant cell tracking error types while \textit{continuously} utilizing the full value range. 
We hope that CHOTA steers algorithmic development more sustainably by incorporating local correctness, global coherence, and lineage information. Addressing all of them in one metric will hopefully lead to advancements that also address all facets of cell tracking.

\section*{Acknowledgements}
\label{sec:acknowledgements}

This work was supported by the Federal Ministry of Education and Research (BMBF), Germany under the AI service center KISSKI (grant no. 01IS22093C), the Lower Saxony Ministry of Science and Culture (MWK) through the program zukunft.niedersachsen of the Volkswagen Foundation, the Deutsche Forschungsgemeinschaft (DFG) under Germany’s Excellence Strategy within the Cluster of Excellence PhoenixD (EXC 2122) and the Ministry of Education, Youth and Sports of the Czech Republic through
the e-INFRA CZ (ID:90254).

\bibliographystyle{splncs04}
\bibliography{main}

\newpage
\appendix
\section{Appendix}
\subsection{CHOTA Implementation}
\label{appendix:chota_implementation}
The CHOTA metric can be implemented efficiently. It is separated into two steps: First, object detections are matched to ground truth annotations. Second, the TPA, FPA and FNA for every True Positive are calculated and summed up. The next section describes both steps in detail and shows that the second step is very efficient in practice. We use the notation of Section 2 and 3.

\paragraph{Matching.} As discussed in Section 2, every tracking metric requires a precedent matching of predicted detections to ground truth annotations. 
Depending on the protocol, the matching can directly be deduced from the intersection of the predicted and ground truth masks (see Section 2.1) or with a subsequent bilinear matching (see Section 2.2). 
Practically, the matching procedure is computationally intensive: Every ground truth and predicted mask needs to be loaded from disk and compared pixel-by-pixel. When neglecting the bilinear mapping, the matching process is basically the calculation of the intersection of instance masks and therefore linear depending on the dataset characteristics, \ie the number of frames and size of the images. 
However, since the matching needs to be performed for every metric, it is not specific for CHOTA.  

\paragraph{Association Scores.} After the matching procedure, We have TP, FP, and FN. 
Furthermore, we have lineage relations $\text{L}^\text{gt}$ and $\text{L}^\text{pr}$ between IDs according to our trajectory definition in Equation (17). The lineage relations can be expressed as a mapping $l(i)=\{i, j_1,\ldots, j_n \}$ that maps a set of $n + 1$ IDs to an ID $i$. Every ID $i$ has at least a relation to itself and to $n$ other IDs depending on the lineage tree. For example, the trajectories in Figure 2b have mappings $l(1)=\{1,3,4\}$, $l(2)=\{2\}$, $l(3)=\{1,3\}$, and $l(4)=\{1,4\}$. 

These sets are used to create a prediction-to-ground-truth mapping accumulator $M\in\mathcal{R}^{(I+1)\times(J+1)}$ with the number of predicted and ground truth IDs $I=|\text{PrID}|$ and $J=|\text{GtID}|$. An element $M_{ij}$ reflects the number of TPs matched to the $i$-th predicted ID and $j$-th ground truth ID. Furthermore, row 0 is used to count FNs, such that $M_{0j}$ contains the number of $|\{ c \in \text{FN} \, |\, \text{gtID}(c) = j \}|$ of ID $j$. Similarly, column 0 aggregates FPs, such that $M_{i0}$ contains the number of $|\{ c \in \text{FP} \, |\, \text{prID}(c) = i \}|$ of the predicted ID $i$. The number of operations to fill the accumulator is linear depending on the number of detections and annotations, \ie there are $|\text{TP}|+|\text{FP}|+|\text{FN}|$ additions required. The resulting accumulator $M$ is usually extremely sparse because most predicted trajectories only overlap with a single or up to a few ground truth trajectories.

\begin{algorithm}[ht]
\caption{Calculate CHOTA}\label{alg:chota}
\begin{algorithmic}[1]
\Require TP, FP, FN, $L^\text{pr}$ $L^\text{gt}$
\State $I,J \gets |\text{PrID}|,|\text{GtID}|$
\State $M \gets \{0\}^{(I+1)\times (J+1)}$
\For{$c\in\text{TP}$} \Comment{Accumulate True Positives} 
    \State $i \gets \text{PrID}(c)$
    \State $j \gets \text{GtID}(c)$
    \State $M_{ij} \gets M_{ij} +1 $
\EndFor
\For{$c\in\text{FP}$} \Comment{Accumulate False Positives}
    \State $i \gets \text{PrID}(c)$
    \State $M_{i0} \gets M_{i0} +1 $
\EndFor
\For{$c\in\text{FN}$} \Comment{Accumulate False Negatives}
    \State $j \gets \text{GrID}(c)$
    \State $M_{0j} \gets M_{0j} +1 $
\EndFor
\State $\mathcal{A}^\sigma \gets 0$   \Comment{Aggregate Matching Scores $\mathcal{A}^\sigma(c)$}
\For{$M_{ij} \in \{M_{ij} \ |\ M_{ij}>0;\ i>0;\ j>0 \}$} 
    \State $\text{TPA}^\sigma \gets \sum_{i'\in l(i)} \sum_{j'\in l(j)} M_{i'j'}$
    \State $\text{FPA}^\sigma \gets  \sum_{i'\in l(i)} \sum_{j'=0}^{J+1} M_{i'j'}  - \text{TPA}^\sigma$
    \State $\text{TPA}^\sigma \gets \sum_{j'\in l(j)} \sum_{i'=0}^{J+1} M_{i'j'}  - \text{TPA}^\sigma$  
    \State $\mathcal{A}^\sigma \gets \mathcal{A}^\sigma + M_{ij} \cdot \frac{\text{TPA}^\sigma}{\text{TPA}^\sigma+\text{FPA}^\sigma+\text{FNA}^\sigma}$   
\EndFor
\State \textbf{return} $\text{CHOTA} \gets \sqrt{\frac{\mathcal{A}^\sigma}{|\text{TP}|+|\text{FP}|+|\text{FN}|}}$
\end{algorithmic}
\end{algorithm}

Now, the matching score $\mathcal{A}^\sigma(c)$ from Equation (18) for a True Positive $c\in \text{TP}$ can be calculated using $M$, $L^\text{gt}$ and $L^\text{pr}$. For an arbitrary $c$ with $\text{PrID}(c)=i$ and $\text{GtID}(c)=j$, the association measures FPA, FNA and TPA can be calculated as follows:
\begin{equation}
    \text{TPA}^\sigma(c)= \sum_{i'\in l(i)} \sum_{j'\in l(j)} M_{i'j'}
\end{equation}
\begin{equation}
    \text{FPA}^\sigma(c)= \sum_{i'\in l(i)} \sum_{j'\not\in l(j)} M_{i'j'} =  \sum_{i'\in l(i)} \sum_{j'=0}^{J+1} M_{i'j'}  -|\text{TPA}^\sigma(c)| 
\end{equation}
\begin{equation}
    \text{FNA}^\sigma(c)= \sum_{j'\in l(j)} \sum_{i'\not\in l(i)} M_{i'j'} =  \sum_{j'\in l(j)} \sum_{i'=0}^{I+1} M_{i'j'} -|\text{TPA}^\sigma(c)|  
\end{equation}
The value $\mathcal{A}^\sigma(c)$ is equal for all $c\in \text{TP}$ that share the same prediction and ground truth ID $\text{PrID}(c)=i$ and $\text{GtID}(c)=j$. Thus, $\mathcal{A}^\sigma(c)$ needs to be calculated only once for all TPs that are represented by an element $M_{ij}$. In other words, the number of to calculate values $\mathcal{A}^\sigma(c)$ equals the number of non-zero elements $M_{ij}>0$ with $i>0$ and $j>0$. Assuming an extremely sparse $M$ with $n$ non-zero elements per ID on average, the entire calculation is quasi-linear depending on the number of trajectory IDs and $n$. However, since the number of trajectory IDs is usually much smaller than the number of detections, the CHOTA calculation complexity is practically negligible compared to the matching procedure. The complete algorithm is shown in Algorithm 1.

\newpage

\foreach \seq/\dataset in {
01/BF-C2DL-HSC,
02/BF-C2DL-HSC,
01/BF-C2DL-MuSC,
02/BF-C2DL-MuSC,
01/DIC-C2DH-HeLa,
02/DIC-C2DH-HeLa,
01/Fluo-C2DL-MSC,
02/Fluo-C2DL-MSC,
01/Fluo-N2DH-GOWT1,
02/Fluo-N2DH-GOWT1,
01/Fluo-N2DL-HeLa,
02/Fluo-N2DL-HeLa,
01/PhC-C2DH-U373,
02/PhC-C2DH-U373,
01/PhC-C2DL-PSC,
02/PhC-C2DL-PSC,
01/Fluo-N2DH-SIM+,
02/Fluo-N2DH-SIM+}
{ 
    \begin{figure}[tb]
    \centering
    
        \begin{subfigure}{.32\textwidth}
            \centering
            \includegraphics[width=1\textwidth]{figures/noise/\dataset _\seq _add_false_negative.pdf}
        \end{subfigure}
        \begin{subfigure}{0.32\textwidth}
            \centering
            \includegraphics[width=1\textwidth]{figures/noise/\dataset _\seq _add_false_positive.pdf}
        \end{subfigure}
        \begin{subfigure}{.32\textwidth}
            \centering
            \includegraphics[width=1\textwidth]{figures/noise/\dataset _\seq _add_idsw.pdf}
        \end{subfigure}
        
        \begin{subfigure}{.32\textwidth}
            \centering
            \includegraphics[width=1\textwidth]{figures/noise/\dataset _\seq _remove_matches.pdf}
        \end{subfigure}
        \begin{subfigure}{0.32\textwidth}
            \centering
            \IfFileExists{figures/noise/\dataset _\seq _remove_mitosis.pdf}{\includegraphics[width=1\textwidth]{figures/noise/\dataset _\seq _remove_mitosis.pdf}}{\includegraphics[width=1\textwidth]{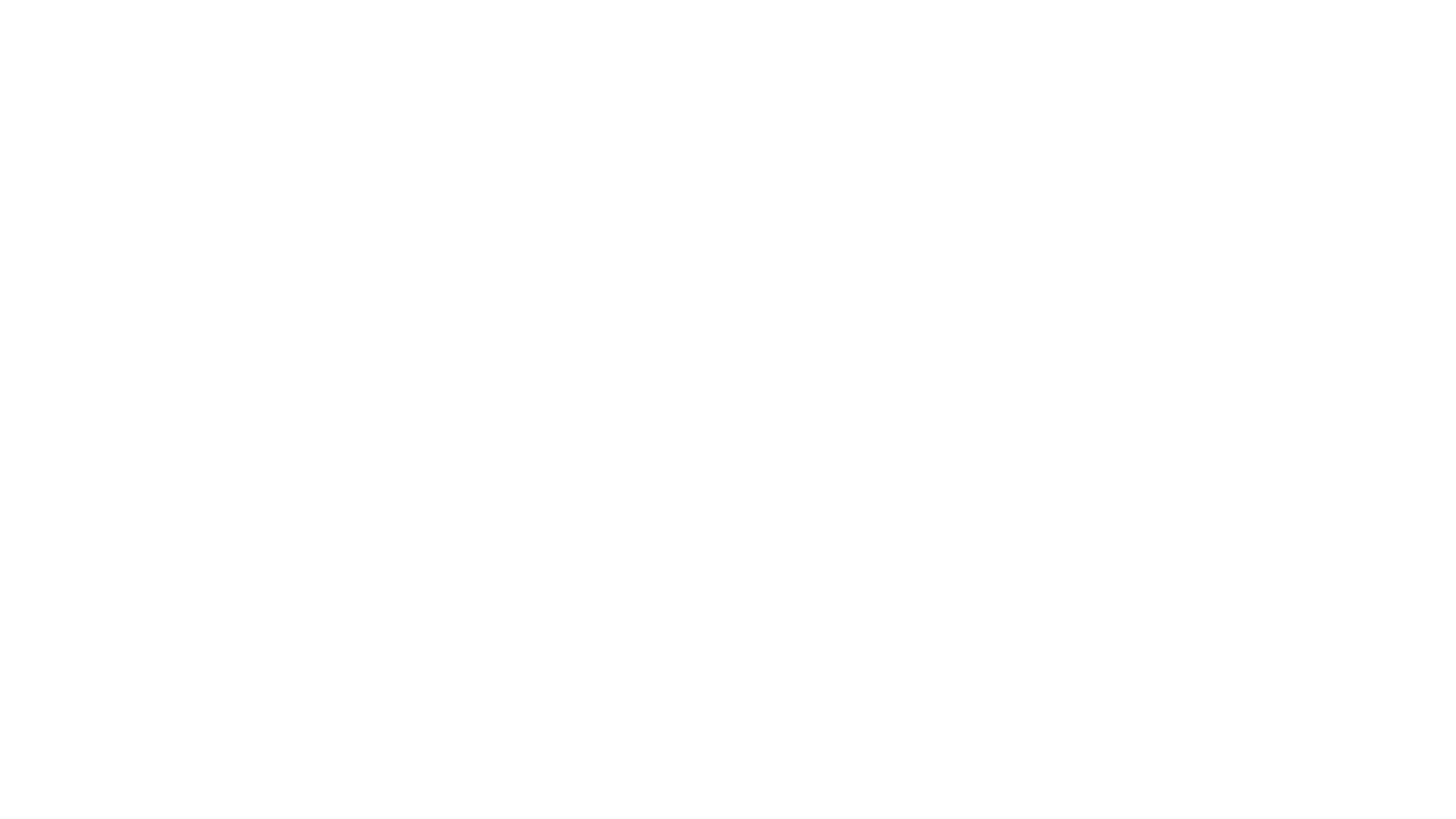}}
            
        \end{subfigure}
        \begin{subfigure}{.32\textwidth}
            \centering
            \includegraphics[width=.4\textwidth]{figures/noise/legend.pdf}
        \end{subfigure}
        
    \caption{Synthetic error induction for Dataset \dataset~(Sequence \seq)}
    \end{figure}
}

\end{document}